\documentclass[journal]{IEEEtran}
\usepackage{multirow,epsfig,fbox,amsfonts,amsmath,multicol,enumitem,algorithm,algorithmic,pifont,graphicx,bm}
\usepackage{booktabs}
\usepackage{bm,enumitem}
\graphicspath{{figs/}}

\usepackage{subfigure}
\usepackage[numbers,sort&compress,comma]{natbib}
\usepackage{hyperref}
\hypersetup{
	colorlinks=true,
	linkcolor=blue
}

\title{OCCDiff: Occupancy Diffusion Model for High-Fidelity 3D Building Reconstruction from Noisy Point Clouds}

\author{Jialu Sui, Rui Liu, Hongsheng Zhang
	\thanks{This study was jointly supported by the National Natural Science Foundation of China. \textit{(Corresponding authors: Hongsheng Zhang)}}
	\thanks{Jialu Sui, Rui Liu, Hongsheng Zhang are with the Department of Geography, Faculty of Social Science, the University of Hong Kong, Hong Kong, China. (e-mail: jialusui99@connect.hku.hk; rhysliu@connect.hku.hk; zhanghs@hku.hk).}
    }
\begin{document}
\maketitle
%%%%%%%%% ABSTRACT
\begin{abstract} 
A major challenge in reconstructing buildings from LiDAR point clouds lies in accurately capturing building surfaces under varying point densities and noise interference. To flexibly gather high-quality 3D profiles of the building in diverse resolution, we propose OCCDiff applying latent diffusion in the occupancy function space. Our OCCDiff combines a latent diffusion process with a function autoencoder architecture to generate continuous occupancy functions evaluable at arbitrary locations. Moreover, a point encoder is proposed to provide condition features to diffusion learning, constraint the final occupancy prediction for occupancy decoder, and insert multi-modal features for latent generation to latent encoder. To further enhance the model performance, a multi-task training strategy is employed, ensuring that the point encoder learns diverse and robust feature representations. Empirical results show that our method generates physically consistent samples with high fidelity to the target distribution and exhibits robustness to noisy data.
\end{abstract}    

%%%%%%%%% BODY TEXT
\section{Introduction}
\label{sec:intro}
3D building reconstruction remains a challenging yet pivotal task, underpinning numerous downstream applications in urban analytics, simulation, and digital twin development \cite{li2021quantifying,cao2021effects}. As urban expansion progresses alongside infrastructural advancement, building-scale digital twins \cite{batty2024digital,clemen2021multi,therias2023city} have been investigated from multiple perspectives, including facades, roofs, and full-structure reconstructions. For outdoor-oriented 3D building modeling \cite{chen2024polygnn,liu2024point2building}, point cloud data, particularly from Airborne LiDAR Scanning (ALS) and Unmanned Aerial Vehicle (UAV), remains the primary data source. However, those point clouds often exhibit sparsity, non-uniform point density, and significant levels of occlusion and noise—factors that complicate the accurate reconstruction of building geometries. Moreover, unlike other 3D reconstruction contexts, building reconstruction frequently necessitates varying levels of geometric detail. In some cases, highly precise 3D information is essential, while in others, computational efficiency and data storage optimization take precedence. Balancing resolution with computational constraints often requires strategic data simplification. As a result, developing robust and adaptive reconstruction methodologies that can effectively balance accuracy, efficiency, and scalability has become a key research priority.

In recent years, many simplified representations for 3D buildings, such as line-based models, have been developed to reduce storage requirements and are widely used in large-scale urban scenarios. However, generating accurate and high-quality 3D building models based on point clouds and other precise 3D representation method remains essential. Previous approaches commonly adopt end-to-end training pipelines, in which an encoder–decoder network first produces a coarse yet complete 3D model \cite{fei2022comprehensive,tesema2023point}, followed by a refinement or upsampling stage to enhance geometric detail. For instance, APC2Mesh \cite{pcc} employed a dynamic multi-scale attention-based network that captures both local and global semantic information to reconstruct complete building geometries, thereby mitigating issues related to occlusion, noise, and sparsity. However, the dependence on encoder–decoder architectures limits model generalization and hinders performance across diverse shape completion tasks. Furthermore, the end-to-end training pipelines with Chamfer Distance (CD) loss can result in ambiguous or smoothed boundaries, which is suboptimal for building reconstruction, where structures are predominantly composed of cuboid elements with sharp edges and corners. Recently, generative models, particularly diffusion models \cite{luo2021diffusion}, have achieved remarkable success across a wide range of applications, including 3D shape generation. In this context, shape completion can be interpreted as a specialized form of conditional 3D shape generation \cite{li2024pccdiff, chen2024learning}, where the model synthesizes complete geometries conditioned on partial observations. DiffComplete \cite{Diffcomplete} represented a notable diffusion-based framework that incorporates a hierarchical feature aggregation mechanism and an occupancy-aware fusion strategy to improve reconstruction fidelity. Despite its effectiveness, the resolution of the generated 3D shapes remains limited by the density of the input point cloud or the granularity of the voxel grid, which constrains the precision of fine geometric details.

As a result, to flexibly generate fine and rough shapes, we introduce a diffusion model within the function-space representation of 3D geometry. Specifically, we design an encoder–decoder-based function autoencoder to learn the complete 3D structure of buildings. Subsequently, a latent diffusion model is introduced to the function space to learn the distribution of the latent representations produced by the latent encoder, enabling more robust and expressive shape generation. To incorporate geometric priors from incomplete observations, we further introduce a point encoder with global-wise attention that directly processes partial point clouds to constrain the predicted occupancy values and provide conditional guidance for diffusion-based learning. This integration enhances the model’s ability to reconstruct accurate and detailed structures. Moreover, we apply the multi-task learning to improve the geometric fidelity of the feature from the point encoder, and the effective conditional feature further improves the robustness of the occ decoder. Extensive experiments conducted on two large-scale building completion datasets demonstrate that our method achieves superior performance, exhibiting high robustness and adjustable computational cost compared to existing approaches.

%In this framework, the latent encoder takes the real occupancy values and corresponding query positions as inputs, while the occupancy decoder (occ decoder) reconstructs the occupancy values to capture spatial continuity.

\begin{itemize}
\item  A novel latent diffusion framework over function space is proposed for achieving 3D reconstruction and completion. It generates continuous function representations from partial input conditions and global diffusion features.

\item A point encoder with global-wise attention is proposed to provide geometric priors for the latent encoder, occ decoder, and diffusion model to reduce the influence of training bias on the final results.

\item The multi-task learning strategy is applied to preserve the geometric fidelity of the feature extracted by the point encoder.

\end{itemize}

\section{Related Work}

\subsection{3D representation}

Various 3D representations have been proposed for the shape completion task, primarily including point-based approaches \cite{pcc,anchorformer} and voxel-based methods \cite{Diffusion-occ,wu2018learning}. In addition, alternative representations such as line-based \cite{EdgeDiff,li2022point2roof,BWFormer}, occupancy-based \cite{occ}, and signed distance fields (SDFs) \cite{Diffcomplete,Diffusionsdf,park2019deepsdf} have also been explored. Each representation offers distinct advantages and limitations in the context of building completion. Point-based representations are highly effective in capturing fine-grained surface details while maintaining computational efficiency. However, because point clouds lack explicit connectivity information, they often introduce ambiguities in shape topology and uneven distribution for generated points, making it challenging to ensure surface continuity during completion. In contrast, voxel-based methods provide strong topological clarity owing to their grid-based structure but face a significant trade-off between spatial resolution and computational cost. Line-based representations are notable for their ability to model the structural skeleton of 3D objects with remarkable compactness, requiring substantially fewer parameters than point clouds or voxels. However, their main limitation lies in their reduced ability to represent complex roof geometries commonly found in building structures. Occupancy-based and SDF-based methods address some of these challenges by providing continuous and implicit representations of 3D geometry. Occupancy values indicate whether a spatial point lies inside or outside the object, while SDFs measure the minimum distance from any point to the object’s surface. SDFs generation is significantly more challenging, requiring higher computational resources and complex network design than predicting occupancy values, as SDF generation involves continuous value regression, whereas occupancy prediction is a straightforward binary classification task. Occupancy-based methods are better suited for representing building geometry, offering lower computational cost and more flexible 3D feature representation—provided when trained with properly labeled data.

\subsection{Encoder-decoder-based network}

The success of convolutional neural networks (CNNs) in 2D image analysis has inspired their extension to 3D data understanding through the development of 3D CNNs. For instance, ODGNet \cite{odrnet} employed a Seed Generation U-Net architecture that utilized multi-level feature extraction and concatenation to enhance the representational capability of seed points. In parallel, Transformer-based architectures \cite{wu2024point,zhao2021point,wang2024pointattn} have significantly advanced 3D data representation and understanding. PoinTr \cite{pointr} adopted a Transformer encoder–decoder framework for point cloud completion. However, despite their effectiveness, encoder–decoder-based architectures and the end-to-end training strategy inherently limit shape generation performance, as they tend to overfit to learned instances and produce overly smooth 3D models. This smoothness, while beneficial for generic object completion, is suboptimal for outdoor building reconstruction, where preserving sharp edges and geometric boundaries is essential for structural accuracy. In the building completion task, most existing studies focus primarily on building reconstruction \cite{li2022point2roof}. BWFormer \cite{BWFormer} was a novel Transformer-based model for building wireframe reconstruction, which addressed this problem by first detecting building corners in 2D and then lifting and connecting them in 3D space. While this approach performs well for buildings with relatively simple structures, achieving denoising and completion simultaneously, it struggles to generalize to more complex scenarios.

\subsection{Diffusion-based network}
Diffusion models are novel generative models that achieve impressive performance on generation tasks especially. Starting from 2D image generation, diffusion shows superior ability in generating high-quality results. In 3D task, initially, diffusion models generate 3D model based on multi-view image generation \cite{diffusion2d,yu2021automatic}. After that, many works \cite{chen2024learning,Sdfusion,lyu2023controllable} started to use diffusion models to make 3D generation directly without 2D image generation. For example, SDF-Diffusion \cite{shim2023diffusion} used denoising diffusion models with continuous 3D representation via SDF. Sc-diff \cite{Sc-diff} combined image-based conditioning through cross-attention and spatial conditioning through the integration of 3D features from captured partial scans. Despite the previous works trying to solve the shape completion task with many 3D representation methods and exquisite model design, the diffusion-based models are suffer from high computational cost and are quite sensitive to the input noise, which limits their generalization ability.

\begin{figure}[t!]
\centering
\includegraphics[width=\linewidth]{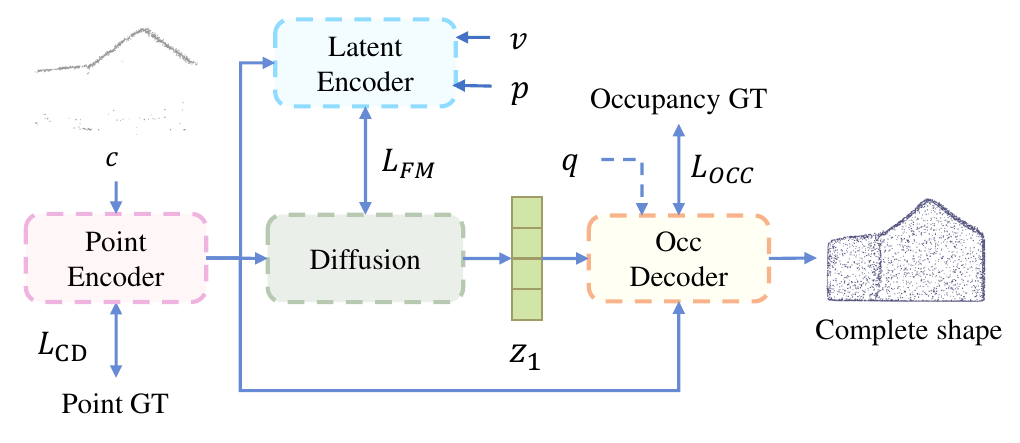}
\caption{The structure of our proposed OCCDiff model.}
\label{fig:occdiff}
\end{figure}

\begin{figure*}[h!]
\centering
\includegraphics[width=\linewidth]{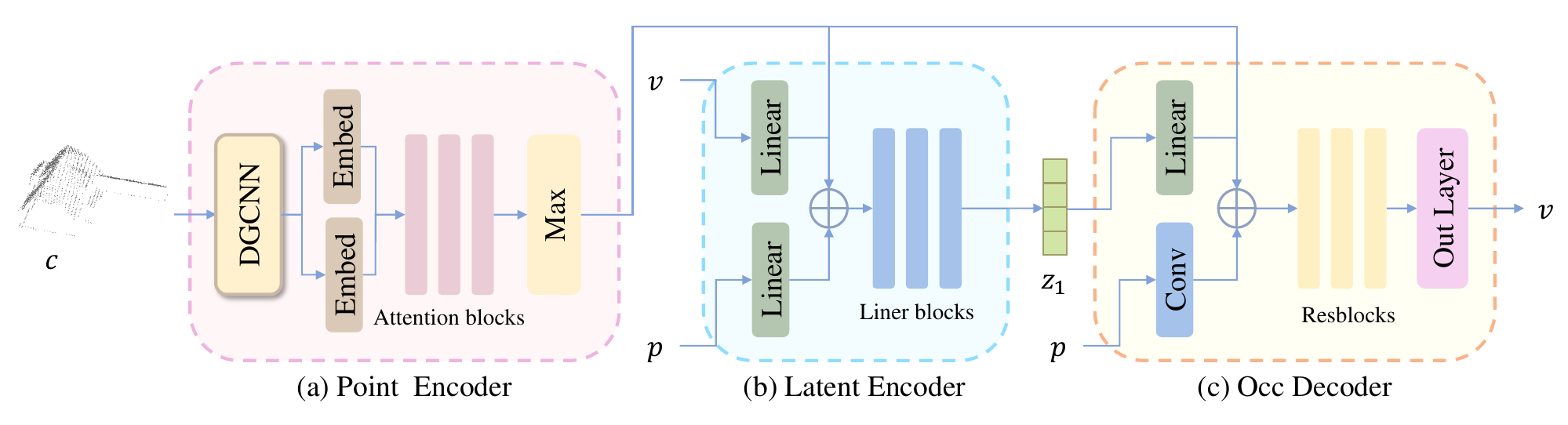}
\caption{The architecture of function autoencoder and point encoder under partial point cloud condition.} 
\label{fig:occ}
\end{figure*}

\section{Method}

\subsection{Preliminaries}
\subsubsection{Flow Matching}
Latent diffusion models \cite{latent} operate in a low-dimensional latent space, which allows efficient processing by abstracting away high-frequency, imperceptible details. However, this approach is often hindered by low inference efficiency, which limits its practicality in many real-time applications. Flow Matching \cite{Flowmatching,Flowmatching2} offers an alternative by learning the data generation process through ordinary differential equations (ODEs). Instead of directly modeling the reverse process as traditional diffusion models, Flow Matching models the flow process of samples between two distributions. These distributions consist of a simple prior distribution, $x_0$, such as standard Gaussian noise, and a complex terminal distribution, $x_1$. The process begins by sampling a point from each distribution and then connecting them through a trajectory. The most common connection method is linear interpolation: $x(t) = (1-t)x_0+tx_1$ $t \in [0,1]$. The objective is to learn a time-dependent velocity field $h_\theta(x,t)$, which describes the instantaneous velocity of each point along the trajectory. The training process aims to match the predicted velocity with the true velocity $\frac{d}{dt}x(t) = x_1-x_0$.
% \begin{equation}\label{eq:qtqt-1}
% 	L_{LDM} := \mathbb{E}_{\psi(c), \epsilon \sim {\cal N}(0,{\bm I}),t} \biggl[ || \epsilon -\epsilon_{\theta}(z_t,t) ||^2_2 \biggr].
% \end{equation}

\subsubsection{Occupancy networks}

Occupancy networks \cite{occ} implicitly represent the 3D surface as the continuous decision boundary of a deep neural network classifier. This approach encodes a continuous 3D representation at infinite resolution, without incurring excessive memory costs. The model generates an occupancy probability at every potential 3D point $s \in \mathbb{R}^3$, defined as

\begin{equation}
 o : \mathbb{R}^3 \to \{0,1\}.
\end{equation}

Given an encoded input condition $y \in \mathcal{X}$, such as the encoded features from an image or point cloud, the occupancy representation can be efficiently parameterized by a neural network $f_\theta$. This network takes a pair $(s, y) \in \mathbb{R}^3 \times \mathcal{X}$ as input, and outputs a real-valued number that represents the probability of occupancy at the position $s$. Formally, this is expressed as:

\begin{equation}
 f_\theta: \mathbb{R}^3 \times \mathcal{X} \to [0,1].
\end{equation}

\subsection{Problem Formulation.}
As shown in Fig.~\ref{fig:occdiff}, our framework aims to generate a complete 3D building model from an occupancy representation based on partial point cloud data. These partial point clouds are often noisy and severely lacking in information, particularly in the facade regions. Specifically, we construct a function autoencoder that encodes the ground truth occupancy values $v \in \mathbb{R}^Q$ and their spatial positions $p \in \mathbb{R}^{Q \times 3}$ to a continuous global feature. Subsequently, a diffusion model is employed over the function space, working as a transformer between the feature from the input of partial point cloud $c$ and the complete continuous global feature $z$. Finally, these completed latent features generate by diffusion are then fed into the pretrained decoder to produce the occupancy result of the query position.

\subsection{Function autoencoder under condition}

As shown in Fig.~\ref{fig:occ}, our function autoencoder comprises two main components: the latent encoder and the occ decoder. Additionally, a point encoder is incorporated to generate conditional features that support the training of the function autoencoder. For the function autoencoder, the latent encoder processes the randomly sampled query positions and their corresponding ground-truth occupancy values to produce global latent features. The occupancy decoder then uses these latent features, together with the query positions, to reconstruct the original occupancy values. Since this occupancy value prediction is essentially a binary classification task, the structure of the latent encoder and the occ decoder are relatively simple and lightweight. As illustrated in Fig.~\ref{fig:occ}, the latent encoder primarily consists of several linear layers for global latent feature extraction, while the decoder utilizes ResBlocks for final occupancy value prediction. The formula of this function autoencoder is shown:
\begin{equation}\label{eq:qtqt-1}
	z_1 = E(p,v,y),
\end{equation}
\begin{equation}\label{eq:qtqt-1}
	v = f_\theta(z_1,p,y),
\end{equation}
where $E$ represents the latent encoder, $y$ is the output of the point encoder.

Our point encoder to extract comprehensive global features from the input partial point cloud, which serves as a condition for the function autoencoder and the diffusion model. As shown in Fig.~\ref{fig:occ}, it utilizes a DGCNN \cite{DGCNN} as the core feature extractor and a series of Transformer layers \cite{anchorformer} to perform global feature extraction. Specifically, DGCNN generates sparse point positions along with their corresponding features, which are embedded separately and then fed into the attention blocks. These attention blocks integrate global and local information to produce the final attention representations. Moreover, to compute the CD loss described in Section~\ref{cdloss}, a simple linear layer is employed to generate a coarse complete point cloud.

The features from the point encoder are input into the latent encoder, decoder, and the diffusion network. For the latent encoder, the features from the point encoder inject multi-modal information, enhancing the latent feature generation. The latent encoder takes occupancy values and their corresponding positions as input, while the point encoder processes the point cloud data. Integrating these two modalities produces a more comprehensive and informative global latent feature. For the occ decoder, the input from the point encoder stabilizes the network's overall performance, since it is hard for the function autoencoder to converge without condition guidance during the training process. Moreover, the point encoder's output also provides a stable, deterministic condition for the occ decoder that reduces reliance on the potentially unstable latent features generated by flow matching. Finally, the feature from the point encoder also serves as a necessary encoded condition to the diffusion network, enabling it to produce more accurate features.

\subsection{Diffusion over function spaces}
Compared with other encoder–decoder-based models, the decoder in the function framework requires both a global continuous latent vector and structured constraint information. The continuous latent vector helps mitigate minor disturbances in the function space, ensuring that the generated geometric surfaces remain smooth and free of cracks or sharp edges. Meanwhile, the structured constraint information enforces geometric consistency with building structures, preventing the generation of irregular or unrealistic shapes. The previous approach \cite{occ} attempted to directly transform partial point clouds into a complete global continuous latent representation with structured constraint information. However, due to the significant discrepancies between the partial and complete features, this direct conversion proved ineffective. To address this, we introduce a diffusion model, specifically leveraging flow matching, which provides a ``buffer zone" that effectively bridges the gap between the incomplete point cloud features and the smooth, continuous global features. Flow matching excels at providing smooth and continuous latent features $z_1$ with strict geometric constraints under partial point cloud conditions.

Moreover, compared to global features derived from other types of 3D data, such as point cloud and voxel for the previous approach \cite{occ,Diffusion-occ}, the smoother, full-rank, and low-noise latent features extracted from the ground truth occupancy values and their positions are particularly well-suited for flow matching. Occupancy-based representations capture richer geometric context by encoding reference values across multiple spatial directions and distances, allowing for meaningful signals along all three coordinate axes. As a result, the encoder’s gradient can be independently unfolded across these multidimensional spatial directions, producing a smoother, full-rank, and low-noise latent space that is ideal for learning and expression through flow matching. Further theoretical comparison analysis about the other features and supporting evidence is provided in the Appendix. As a result, flow matching is particularly suitable for this task because it naturally produces smooth and continuous global features. Moreover, the global features derived from ground-truth occupancy values and their corresponding positions align well with the flow-matching learning process.

During the training of the function autoencoder, the latent encoder in the function autoencoder transforms discrete position and occupancy values into continuous geometric features $z_1$ with structured constraint information, which serve as global conditional features to guide the decoder in generating occupancy values. After that, flow matching learns a continuous vector field that guides samples from an initial simple distribution $z_0$ to our target distribution $z_1$ through a smooth and continuous transformation process. During inference, we initialize $z_0$ by sampling from a standard normal distribution and iteratively refine it through $N$ denoising steps to obtain $z_1$, which is then fed into the occ decoder to reconstruct the complete 3D geometry. The update process of $z_{t+\Delta t}$ can be formulated as:

\begin{equation}
z_{t+\Delta t} = z_t + \Delta t\cdot h_\theta(z_t,t),
\end{equation}  
\begin{equation}
h_\theta(z_t,t) = z'_1-z'_0,
\end{equation}  
where $\Delta t = \frac{1}{N}$, $z'_1$ and $z'_0$ are predicted result to distinguish them from the real $z_1$ and $z_0$.

\begin{figure}[t!]
\centering
\includegraphics[width=0.85\linewidth]{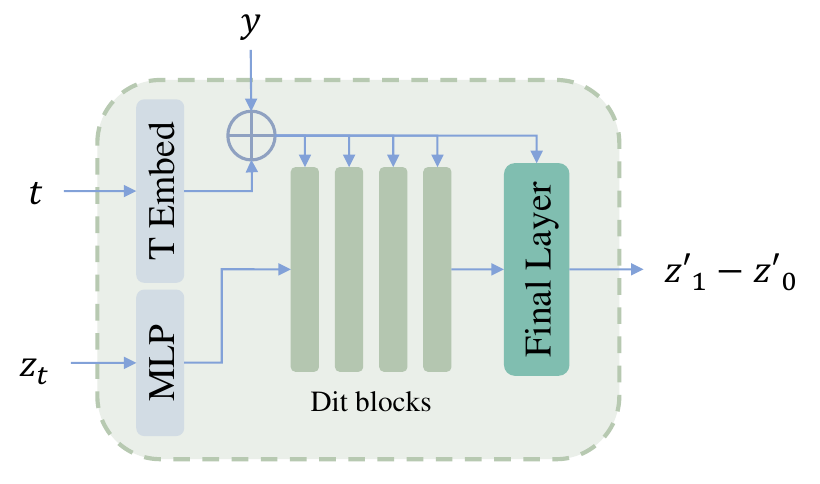}
\caption{The structure of the network in the diffusion model. Since our input feature $z_t$ is one-dimensional feature, we use a simple MLP as input embedding and add learnable position embedding.}
\label{fig:dit}
\end{figure}

For the diffusion model network, we adopt DiT \cite{Dit} as the backbone in our experiments. However, unlike image-based inputs with dimensions $W \times H \times C$, where $C$ denotes the number of channels, our latent representation $z_1$ is a one-dimensional feature vector of size $B \times C$. Consequently, we modify the patch embedding module of DiT by replacing it with a simpler linear embedding layer and incorporating learnable positional embeddings, as illustrated in Fig.~\ref{fig:dit}.

\begin{figure}[t]
\centering
\includegraphics[width=\linewidth]{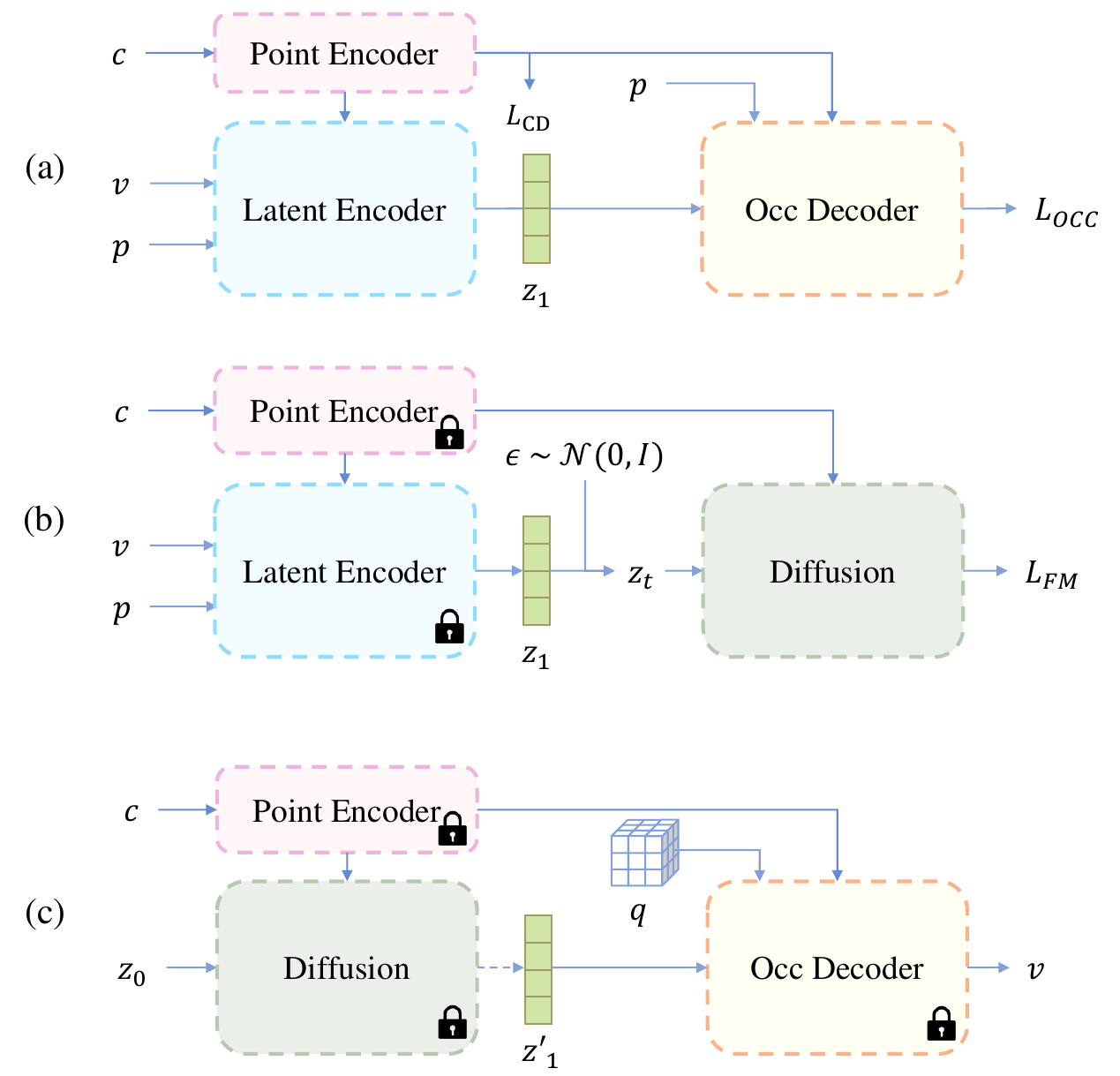}
\caption{The process of training and inference. Point Encoder is omitted in each figure. (a) Training of function autoencoder. (b) Training of  the diffusion model. (c) Inference. }
\label{fig:process}
\end{figure}
   
\subsection{Multi-task Training and Inference}\label{cdloss}

The training process of our model is divided into two main stages: (a) function autoencoder and point encoder, and (b) diffusion model, as shown in Fig.~\ref{fig:process} (a) and (b). We apply the multi-task training strategy in the first stage (a), using a combination of occupancy loss and CD loss. The occupancy loss ensures accurate encoding and decoding within the function autoencoder, while the CD loss enforces geometric consistency for the coarse point cloud generated by the point encoder. The introduction of CD loss explicitly strengthens the learning of the point encoder, ensuring that it not only stabilizes the training of the function autoencoder but also provides rich and informative conditional features for the diffusion model training. The overall loss function follows the formulation proposed in \cite{occ}, and is defined as:

\begin{equation}\label{eq:qtqt-1}
	L_{o} :=L_{occ}+\eta * L_{CD},
\end{equation}
\begin{equation}\label{eq:qtqt-1}
	L_{occ} :=\sum_{j=0}^{K} {\cal L}(f_\theta(p_j,c),o_{j}),
\end{equation}
\begin{equation}\label{eq:qtqt-1}
	L_{CD} := CD_{L2}(q,g).
\end{equation}

where $\eta$ denotes the weight parameters, $q$ and $g$ represent the predicted sparse point cloud and the ground-truth point cloud sampled from the mesh, respectively.

In process (b), we employ a diffusion model to learn the global 3D structural feature produced by the latent encoder, conditioned on the feature embeddings provided by the point encoder. During this stage, the parameters of the latent encoder and point encoder are frozen. To achieve stable and efficient training, we adopt a flow matching strategy for the diffusion model, enabling it to learn the feature distribution of the complete building shapes. The corresponding loss function is defined as:
 
\begin{equation}\label{eq:qtqt-1}
	L_{FM} := \mathbb{E}_{z_0,z_1,t} \biggl[ || (z_1-z_0) - h_\theta(z_t,t,c) ||^2_2 \biggr].
\end{equation}
where $h_\theta(z_t,t,c)$ represents the predicted velocity by the neural network.

In Fig.~\ref{fig:process} (c), during inference, a randomly sampled latent vector $z_0 \sim {\cal N}(0,{\bm I})$ is fed into the diffusion model to generate $z_1$. Simultaneously, the partial point cloud is processed by the point encoder to obtain the feature $c$. Finally, $c$, $z'_1$, and the query positions are input to the occ decoder to reconstruct the complete 3D shape of the building. The conversion from occupancy values to the final mesh is performed using Multiresolution IsoSurface Extraction (MISE) \cite{occ}.

\begin{table*}
	\centering
	\setlength{\tabcolsep}{4mm}{
		\begin{tabular}{c|ccc|ccc}
			\hline
                \multirow{2}{1cm}{\textbf{Method}}&
			\multicolumn{3}{c|}{\textbf{Building3D}}&\multicolumn{3}{c}{ \textbf{Building-PCC}}\cr
			    &\textbf{$CD_{L2}$} $\downarrow$ &\textbf{$CD_{L1}$} $\downarrow$&\textbf{$F-Score$ $\uparrow$}&\textbf{$CD_{L2}$} $\downarrow$ &\textbf{$CD_{L1}$} $\downarrow$&\textbf{$F-Score$ $\uparrow$}\cr
			\hline
                Pointr (2021) \cite{pointr}               & 4.89 & 79.86 & 0.3341 & 4.06 & 61.13 & 0.4253  \\
			APC2Mesh (2024) \cite{pcc}                         & 3.90 & 162.61& 0.4890 & 2.93 & 31.32 & 0.6421 \\
                AnchorFormer (2023) \cite{anchorformer}   & 3.41 & 50.43 & 0.5854 & \textbf{2.78} & 26.83 & 0.6936  \\
                ODGNet (2024) \cite{odrnet}               & 3.24 & 44.08 & 0.5404 & 2.97 & 29.21 & 0.4996 \\
                Onet (2019) \cite{occ}                    & 3.58 & 37.17 & 0.6547 & 2.99 & 22.30 & 0.6546   \\
                ProxyFormer (2023) \cite{proxyformer}     & 3.75 & 53.74 & 0.4982 & 3.29 & 36.47 & 0.5479  \\
			\hline
			OCCDiff & \textbf{3.20} & \textbf{28.86} & \textbf{0.6899} & 2.86 & \textbf{18.84} & \textbf{0.6952}   \\
			\hline
	\end{tabular}}
        \caption{Performance comparison in terms of $CD_{L2}$, $CD_{L1}$, and $F-Score$ on the Building3D and BuildingPCC dataset.}
        \label{tab:building}
\end{table*}

\begin{figure*}[h!]
\centering
\includegraphics[width=\linewidth]{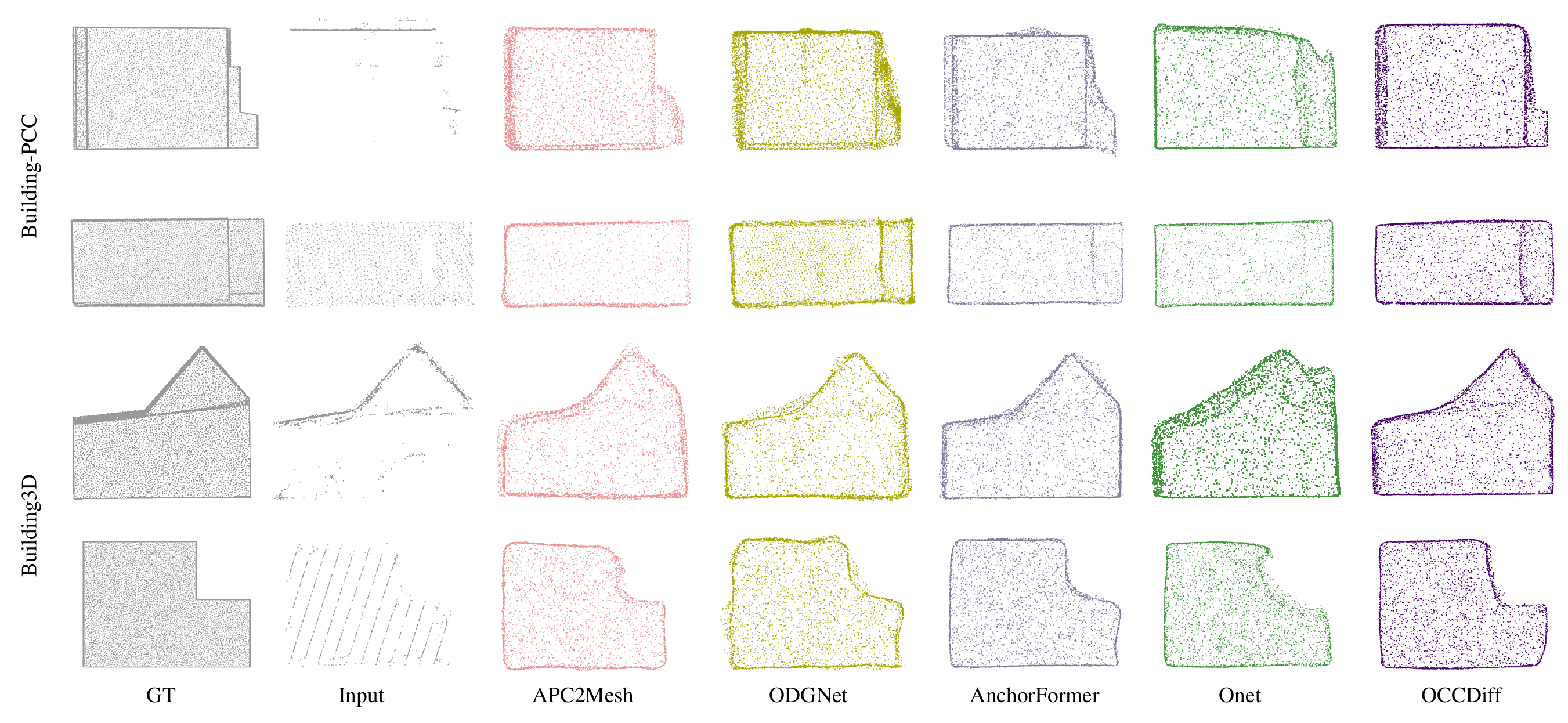}
\caption{Four visual examples of shape completion results by different approaches on the Building3D and Building-PCC dataset. All point clouds are average sampling from generated mesh. Different colors denote the point clouds reconstructed by different approaches.} 
\label{fig:visual}
\end{figure*}

\section{Experiments}
\subsection{Experimental Settings}

\textbf{Datasets.}
We conduct experiments on Building3D \cite{pcc} and Building-PCC \cite{BuildingPCC}. Building3D comprises 15,890 labeled building objects across six cities, while Building-PCC, extracted from The Hague and Rotterdam in the Netherlands, contains approximately 50,000 building models. Both datasets provide partial point cloud data along with corresponding ground-truth 3D models. Notably, each building in Building-PCC includes two types of partial point clouds; during training and testing, we randomly select one of these as input for the model.

\textbf{Implementation Details.}
Our experiments are implemented using the PyTorch framework and conducted on a single NVIDIA GeForce RTX 4090 GPU with 24 GB of memory. The model is trained using the Adam optimizer with a learning rate of ${10}^{-4}$ and batch size 64. For efficient computation, the DiT backbone is configured with a depth of 12, hidden size of 512, patch size of 1, and 16 attention heads. The dimensions of the latent vectors $z$ and $c$ are set to 128 and 512, respectively, while the number of query positions for the function autoencoder is set to 1000. The weight parameter $\eta$ for $L_{occ}$ is set to 1000 and the resolution for our OCCDiff is set to 80. The results of these point cloud completion methods are obtained after surface reconstruction. We sample the point clouds from the reconstructed mesh to make fair comparison across all methods.

\textbf{Evaluation Metrics.}
We employ the L1/L2 Chamfer Distance (CD) $\times 10^{-3}$ and the F-Score as the evaluation metrics for the Building3D and Building-PCC datasets.
%为什么不和diffusion模型比？
%为什么不和直接生成mesh的比？
\subsection{Comparisons with State-of-the-Art Methods}

\textbf{Evaluation in Building3D and Building-PCC.}
We list the quantitative performance of several methods on Building3D and Building-PCC in Table~\ref{tab:building}. It can be seen that OCCDiff achieves better performance in most of the listed metrics and especially achieves the lowest $CD_{L1}$, $CD_{L1}$, and highest $F-Score$. OCCDiff achieves 0.3 $CD_{L2}$, 9 $CD_{L1}$ and 0.03 F1 gain in Building3D and 0.1 $CD_{L2}$, 4 $CD_{L1}$ and 0.04 F1 gain in Building-PCC over our baseline Onet. It proves that the objects completed by OCCDiff have the closest density distribution to ground truth, while there are minor local positioning inaccuracies in a small number of points. The strong performance of $CD_{L1}$ compared with the other methods indicates a high alignment between the completed point cloud and the ground-truth point cloud in terms of overall shape, with strong macro-contour consistency. It also shows low sensitivity to local extreme deviations, enabling stable restoration of the building’s overall structure. For example, though AnchoerFormer achieves better $CD_{L2}$ in Building-PCC, $CD_{L1}$ of OCCDiff is much higher, demonstrating the higher deviations on the overall structure. On the other hand, the excellent $F-Score$ means the algorithm achieves a good balance between point cloud coverage of key structures and redundancy control. It not only accurately covers the core regions of the ground-truth point cloud but also effectively avoids the generation of redundant points, ensuring the validity and practicality of the completion results.

Fig.~\ref{fig:visual} presents a visual comparison among four high-performing different models after surface reconstruction and point sampling. OCCDiff produces high-quality building shapes with smoother wall surfaces, sharper corners, and fewer noise artifacts. For instance, in the last row of Fig.~\ref{fig:visual}, the central corner generated by OCCDiff is noticeably clearer and more defined than those produced by other methods. Although point cloud–based approaches can represent complex roof structures, their results are often highly discrete, making surface reconstruction difficult. The reconstructed meshes frequently exhibit excessive distortion or over-smoothing due to parameter sensitivity, which limits their effectiveness for accurate building completion and reconstruction. In contrast, OCCDiff directly generates complete mesh results, allowing flexible adjustment of output quality based on specific requirements, while also achieving significantly faster surface reconstruction.

\subsection{Analysis of OCCDiff}

\begin{figure}[t]
\centering
\includegraphics[width=\linewidth]{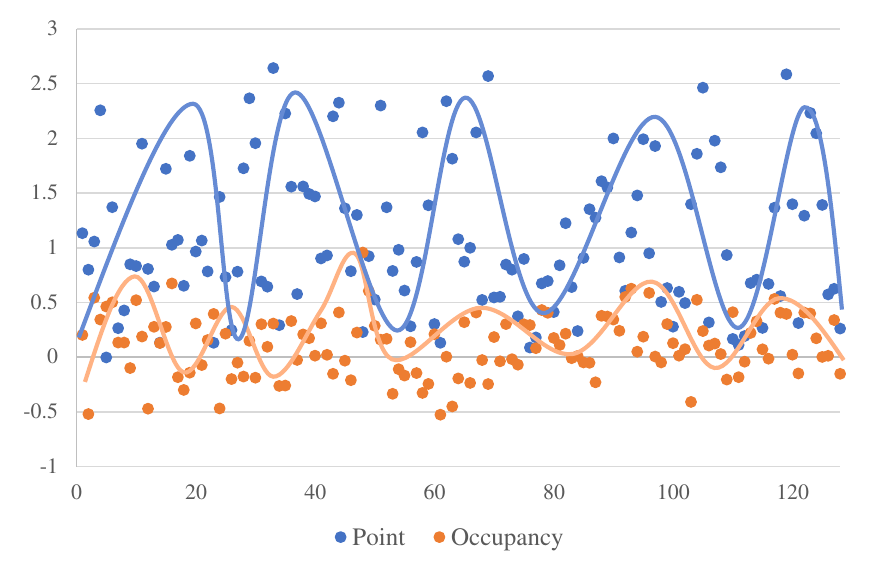}
\caption{The distribution of features from ground truth point clouds and occupancy values after encoding.}
\label{fig:feature}
\end{figure}

\begin{table}[t]
	\centering
	\caption{Comparison of point and occupancy latent feature on Building3D.}
	\setlength{\tabcolsep}{0.5mm}{
		\begin{tabular}{ccccc|c}
			\hline		
                \centering\textbf{Type}&\textbf{Source}& \textbf{$CD_{L2}$} &\textbf{$CD_{L1}$}&\textbf{$F-Score$} & Memory (MB) \\
                \hline
               Point  & Latent  & 2.67 & 16.51 & 0.7681 & 13889\\
                         & Diffusion& 4.16 & 41.53 & 0.4749 &5748\\
                Occupancy&Latent & 2.83 & 18.97 & 0.6721 & 6446\\
               & Diffusion& 3.20 & 28.86 & 0.6899 &5748\\
		\hline
	\end{tabular}}
	\label{tab:Ablation1}
\end{table}

\textbf{Comparison of different latent feature.} Table~\ref{tab:Ablation1} presents the comparison between point-based and occupancy-based latent features on the Building3D dataset. The point-based latent features are generated by a point cloud encoder that processes the ground-truth point cloud data. This encoder is trained in our function autoencoder in place of the latent encoder, enabling it to learn latent representations directly from point clouds. The Latent source represents the output produced by the function autoencoder, whereas the Diffusion source represents the final reconstruction result generated by our OCCDiff framework. As shown, although the point-based features achieve slightly better results in the Latent stage, they also result in higher computational cost. In our experiments, only 2,096 points are sampled for training, yet the method still incurs substantial memory consumption of 13889 MB. Moreover, in the Diffusion stage, the worse results from point clouds show that the features from point clouds are less suitable for diffusion model training compared to the features from occupancy value. The diffusion model trained on occupancy-based features achieves about 1, 13 and 0.2 improvement in $CD_{L2}$, $CD_{L1}$,and $F-Score$ than that trained on point-based features. Point cloud inputs are inherently non-uniform and surface-distributed, lacking balanced sampling in empty space. Consequently, the diffusion network becomes more sensitive to these asymmetric geometric feature distributions, leading to unstable and less coherent latent representations. As illustrated in Fig.~\ref{fig:feature}, the point-based features are notably more dispersed than those derived from occupancy representations.

\begin{table}[t]
	\centering
	\caption{Ablation study on CD loss on Building3D. CD means CD loss function.}
	\setlength{\tabcolsep}{1.3mm}{
		\begin{tabular}{cccccc}
			\hline		
                \centering\textbf{Source}&\textbf{CD}& \textbf{$CD_{L2}$} &\textbf{$CD_{L1}$}&\textbf{$F-Score$} \\
                \hline
                %&&& 3.58 & 37.17 & 0.6547\\
                Latent&& 2.76 & 26.20 & 0.7618\\
               Latent&\checkmark& 2.83 & 18.97 & 0.6721\\
                Diffusion&& 3.47 & 32.34 & 0.6347\\
			Diffusion&\checkmark& 3.20 & 28.86  & 0.6899\\
		\hline
	\end{tabular}}
	\label{tab:Ablation2}
\end{table}

\textbf{Ablation study.} The ablation study on the CD loss is presented in Table~\ref{tab:Ablation2}. The first two rows report results from the function autoencoder, which achieves higher overall performance. The introduction of CD loss slightly reduces the function autoencoder's performance, particularly in terms of $CD_{L2}$ and $F-Score$, indicating that the CD loss on the point encoder may negatively affect function autoencoder training. However, in the diffusion stage, the model conditioned on the point encoder with CD loss demonstrates significant improvement. This improvement arises because, without the CD loss, the point encoder functions only as an auxiliary feature provider for the functional autoencoder, while the ground-truth occupancy values and positions already contain most of the information needed to generate $z_1$. As a result, the point encoder cannot capture the building’s complete geometric structure, which limits the quality of features it provides to the diffusion model and constrains the model’s overall performance.

\begin{table}[t]
	\centering
	\caption{Ablation study on the point encoder feature for latent encoder and occ decoder.}
	\setlength{\tabcolsep}{1mm}{
		\begin{tabular}{cccccc}
			\hline		
                \centering\textbf{Source}&\textbf{Encoder}&\textbf{Decoder}& \textbf{$CD_{L2}$} &\textbf{$CD_{L1}$}&\textbf{$F-Score$} \\
                \hline
               Latent& && 6.72 & 101.68 & 0.4351\\
               Latent&\checkmark& & 6.23 & 126.50 & 0.4723\\
               Latent&&\checkmark & 2.72 & 17.67 & 0.7605\\
               Latent&\checkmark&\checkmark & 2.83 & 18.97 & 0.6721\\
               Diffusion& &\checkmark& 3.53 & 32.62 & 0.6044\\
               Diffusion& \checkmark&\checkmark& 3.20 & 28.86 & 0.6899\\
		\hline
	\end{tabular}}
	\label{tab:Ablation3}
\end{table}

Table~\ref{tab:Ablation3} presents the ablation study on the impact of the point encoder features for both the latent encoder and the occ decoder. The first two rows show the results of the function autoencoder without conditional features provided to the decoder. As illustrated, the occ decoder fails to capture geometric variations across samples in the absence of such conditioning. Consequently, the gradient signals become unstable, and the training process fails to converge. Due to this severe degradation, further experiments without conditional features for the occ decoder are not conducted. For the encoder-side conditioning as shown in the third and fourth rows, while incorporating the point encoder feature slightly limits the function autoencoder’s performance in the first stage, it significantly benefits the diffusion model in the subsequent stage. The multimodal information integrated of the point cloud information and the occupancy value information within the latent encoder facilitates more effective learning of the global 3D shape distribution, leading to improvements of approximately 0.3, 4, and 0.08 in $CD_{L2}$, $CD_{L1}$, and $F$-Score, respectively.

\section{Conclusion}
We propose OCCDiff, a diffusion model in function space, designed to flexibly generate both fine-grained and coarse complete building shapes. Firstly, we design a function autoencoder to compress and reconstruct the complete 3D structural information, yielding compact latent features. A diffusion model is then employed to learn the distribution of these functional latent representations derived from complete 3D models. Flow matching is particularly suitable for this framework because it naturally produces smooth, continuous global features, as the latent encoder's global features provide an ideal representation for flow matching to learn from. To incorporate geometric priors from incomplete observations, we introduce a point encoder with global attention, which directly processes partial point clouds to constrain the predicted occupancy values and provide conditional guidance for the diffusion model. Additionally, multi-task learning is applied during the training of the function autoencoder and point encoder to preserve geometric fidelity in the features extracted by the point encoder. Extensive experiments on two large-scale building completion datasets demonstrate that OCCDiff achieves state-of-the-art reconstruction accuracy and robust generalization, effectively balancing detail preservation and computational efficiency.

%%%%%%%%% REFERENCES
\bibliographystyle{IEEEtranN}
\bibliography{reference}

\end{document}